\documentclass{article} 
\usepackage{iclr2027_conference,times}
\iclrfinalcopy

\usepackage{amsmath,amsfonts,bm}

\def\eqref#1{equation~\ref{#1}}

\def\1{\bm{1}}

\DeclareMathAlphabet{\mathsfit}{\encodingdefault}{\sfdefault}{m}{sl}
\SetMathAlphabet{\mathsfit}{bold}{\encodingdefault}{\sfdefault}{bx}{n}

\usepackage{hyperref}
\usepackage{url}
\usepackage{booktabs}
\usepackage{hyperref}
\usepackage[utf8]{inputenc}
\usepackage{xcolor}
\usepackage[normalem]{ulem}
\usepackage{wrapfig}
\usepackage{subcaption}
\usepackage{times}
\usepackage{latexsym}
\usepackage{multirow}
\usepackage{tabularx}
\usepackage{pifont}
\usepackage{xcolor}
\newcommand{\sota}[1]{\textcolor{black}{\textbf{#1}}}
\usepackage{tcolorbox}
\usepackage{xcolor}
\usepackage{multirow}      
\usepackage{booktabs}      
\usepackage{adjustbox}     
\usepackage{caption}       
\usepackage{geometry}      
\usepackage{amssymb}
\usepackage[T1]{fontenc}
\usepackage[utf8]{inputenc}
\usepackage{textcomp}
\usepackage{microtype}
\usepackage{inconsolata}
\usepackage{seqsplit}        
\usepackage{xcolor}

\usepackage{booktabs}
\usepackage{wrapfig}
\usepackage{xcolor}
\usepackage{graphicx}
\usepackage{framed}
\usepackage{pifont}
\usepackage{enumitem}
\definecolor{shadecolor}{rgb}{0.92,0.92,0.92}
\usepackage{tikz}
\usetikzlibrary{shapes.geometric, arrows.meta, positioning}

\tikzstyle{box} = [rectangle, rounded corners, minimum width=3.2cm, minimum height=1cm, text centered, draw=black, fill=gray!10]
\tikzstyle{process} = [rectangle, minimum width=3.2cm, minimum height=1cm, text centered, draw=black, fill=blue!10]
\tikzstyle{decision} = [diamond, draw=black, fill=yellow!30, minimum size=1.2cm, text centered, inner sep=0pt, aspect=2]
\tikzstyle{arrow} = [thick, ->, >=stealth]
\PassOptionsToPackage{table}{xcolor}

\title{MoEGen: Mixture-of-Experts for Instance Adaptive LoRA Generation}

\author{
\parbox{0.98\textwidth}{
\centering
\vspace*{0.3em}
Yiming Zeng$^{1,\dagger}$ \quad
Lei Lu$^{2,\dagger}$ \quad
Zexin Li$^{3,*}$ \quad
Zhuochun Li$^{4}$ \quad
Dehai Min$^{5}$ \\[0.3em]
Shuoqiu Li$^{6}$ \quad
Shuyi Liao$^{1}$ \quad
Xidong Wu$^{7}$ \quad
Zeyu Zhang$^{8}$ \\[0.3em]
Minmei Wang$^{1}$ \quad
Yu Zhao$^{9}$ \quad
Tingting Yu$^{1,*}$ \quad
Shangqian Gao$^{6,*}$ \\[0.8em]
\normalfont\small
\normalfont\small
\footnotesize
$^{\dagger}$Equal contribution. \quad
$^{*}$Corresponding authors.
}
}

\begin{document}

\maketitle
\lhead{Preprint}
\begin{abstract}
Parameter-efficient fine-tuning (PEFT) enables efficient adaptation of
large language models, but existing MoE-based PEFT methods typically improve capacity by storing multiple full LoRA experts, causing adapter storage to grow linearly with the number of experts and restricting adaptation to a fixed expert pool.
We ask whether MoE-based PEFT can produce instance-specific adaptations without explicitly storing a separate LoRA module for each expert. To address this gap, we propose \textbf{MoEGen}, an adaptation framework that shifts MoE-based PEFT from expert selection to expert-conditioned parameter generation. Instead of storing each expert as a full LoRA adapter, MoEGen represents each expert as a small learnable vector, termed an expert code. It routes each input over these vectors and uses their weighted combination to condition a lightweight hypernetwork that generates input-specific low-rank updates. This design decouples expert capacity from adapter storage while enabling instance-conditioned adaptation. Experiments on eight commonsense reasoning benchmarks show consistent improvements over strong static and MoE-based PEFT baselines across three backbones. MoEGen also performs strongly in joint medical and legal-domain adaptation.
\end{abstract}

\section{Introduction}

Modern pretrained LLMs demonstrate remarkable performance across a wide
spectrum of natural language processing tasks~\cite{fedus2022switchtransformersscalingtrillion,
touvron2023llamaopenefficientfoundation,
chowdhery2022palmscalinglanguagemodeling,olmo20252olmo2furious}. However,
for challenging domain-specific downstream tasks, e.g., code generation and mathematical
reasoning~\cite{roziere2024codellamaopenfoundation,
lewkowycz2022solvingquantitativereasoningproblems}, model adaptation is a standard way to further improve task performance.~\cite{openai2024gpt4technicalreport,
touvron2023llamaopenefficientfoundation}. While full supervised fine-tuning is an effective model adaption method, it is computationally expensive and requires storing separate task-specific model copies, making it less practical in resource-constrained settings~\cite{aghajanyan-etal-2021-intrinsic,
lialin2024scalingscaleupguide}.

Parameter-efficient fine-tuning (PEFT) methods provide practical alternatives
for model adaptation by updating only a small subset of model
parameters~\cite{li2023loftqlorafinetuningawarequantizationlarge,
tian2024hydraloraasymmetricloraarchitecture,
hu2022lora,
liu2024doraweightdecomposedlowrankadaptation}.
LoRA~\cite{hu2022lora}, for example, learns low-rank weight updates while
keeping the backbone parameters frozen. Recent MoE-LoRA methods further expand
adaptation capacity by selecting or combining multiple LoRA experts through a
learned router~\cite{dou-etal-2024-loramoe,
liu2024moemeetsllmsparameter,gao-etal-2025-mola}.
However, because each expert stores a complete set of low-rank parameters,
adapter storage grows linearly with the size of the expert pool, making a large
and diverse adaptation space costly to maintain.

To address these limitations, we propose \textbf{MoEGen}, a lightweight
plug-and-play adaptation framework that synthesizes input-specific LoRA
parameters from compact latent expert codes. Instead of representing each
expert as a full LoRA module, MoEGen encodes experts as small learnable
vectors, termed expert codes, which are independent of the targeted module
dimensions. Prompt-conditioned routing selects multiple expert codes in
this low-dimensional latent space, and a shared hypernetwork generates
their corresponding LoRA parameters, which are then combined to form
input-specific updates. This decouples expert capacity from explicit
backbone-specific adapter storage and and enables flexible input-conditioned adaptation through combinations
of generated LoRA parameters.

Overall, our contributions are as follows:

\begin{itemize}[leftmargin=10pt]
    \item \textbf{Lightweight expert parameterization.}
    MoEGen represents each expert as a low-dimensional learnable code
    rather than a complete LoRA adapter, substantially reducing the parameter and
    storage costs of maintaining multiple experts.

    \item \textbf{Dynamic input-conditioned LoRA generation.}
MoEGen dynamically generates an input-specific low-rank matrix for each adapted component. This enables more targeted adaptation by tailoring the
low-rank update to the semantic characteristics of each input.

    \item \textbf{Consistent empirical performance.}
    MoEGen improves over the strongest competing baselines by
    0.6--1.1 points across eight commonsense reasoning benchmarks while
    updating only 0.44--0.52\% of model parameters. It further achieves
    improvements of 4.6--6.2 points on the cross-domain joint benchmark.
\end{itemize} 

\begin{figure*}
    \centering
    \includegraphics[width=0.9\textwidth]{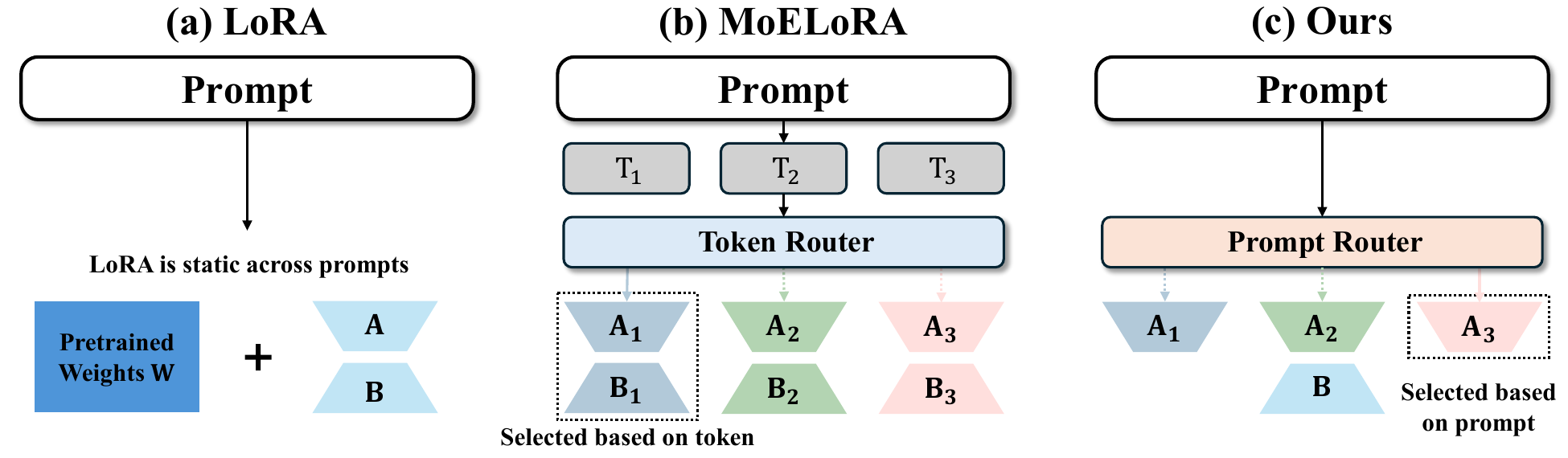}
    \caption{
Comparison between traditional MoE and MoEGen. MoEGen replaces full
feed-forward experts with compact style embeddings and a shared hypernetwork
for input-specific LoRA generation. 
}
    \label{fig:arch}
    \vspace{-5mm}
\end{figure*}

\section{Related Work}
\subsection{Parameter-Efficient Fine-Tuning}

Parameter-efficient fine-tuning adapts pretrained models with few trainable parameters.
Adapters insert bottleneck modules~\citep{houlsby2019parameterefficienttransferlearningnlp,pfeiffer2021adapterfusionnondestructivetaskcomposition}, while prefix-tuning adds learnable prefixes~\citep{li2021prefixtuningoptimizingcontinuousprompts}, with potential latency and expressiveness trade-offs~\citep{han2024parameterefficient}.
LoRA learns low-rank weight updates~\citep{hu2022lora}, which AdaLoRA extends through adaptive rank allocation~\citep{zhang2023adaloraadaptivebudgetallocation}, DoRA through weight decomposition~\citep{liu2024doraweightdecomposedlowrankadaptation}, and LoftQ through quantization-aware initialization~\citep{li2023loftqlorafinetuningawarequantizationlarge}.
Alternatively, SpIEL~\citep{ansell2024scalingsparsefinetuninglarge} and SMT~\citep{he2025smt} update selected parameters or sub-matrices.
These approaches typically share learned adaptations across inputs, whereas MoEGen generates input-conditioned updates.

\subsection{Mixture-of-Experts}

Mixture-of-Experts increases model capacity through sparse expert routing~\citep{shazeer2017outrageouslylargeneuralnetworks}, as demonstrated by Switch Transformer~\citep{fedus2022switchtransformersscalingtrillion} and GLaM~\citep{du2022glamefficientscalinglanguage}.
Extending this paradigm to PEFT, MoELoRA~\citep{liu2024moemeetsllmsparameter} and LoRAMoE~\citep{dou-etal-2024-loramoe} route among multiple LoRA experts, while HydraLoRA introduces asymmetric expert structures~\citep{tian2024hydraloraasymmetricloraarchitecture}.
To make expert allocation more flexible, LD-MoLE uses differentiable routing to adapt the number of activated experts for each token across layers~\citep{zhuang2026ld}.
Despite these advances, maintaining expert-specific adapter parameters ties expert-pool expansion to storage growth.
In contrast, MoEGen routes over compact expert codes and generates input-conditioned LoRA parameters through a shared hypernetwork as shown in Fig.~\ref{fig:arch}.
\subsection{Hypernetworks}

Hypernetworks generate weights for another network \citep{ha2016hypernetworks}.
For efficient adaptation, HyperFormer generates task-conditioned adapters to share knowledge across tasks \citep{karimi-mahabadi-etal-2021-parameter}, while Hyperdecoders extends conditioning to individual inputs \citep{ivison-peters-2022-hyperdecoders}.
This paradigm also supports dialect adaptation through HyperLoRA \citep{xiao2023task}, request-specific text editing through HyperEdit \citep{zeng-etal-2026-hyperedit}, and context internalization through Doc-to-LoRA \citep{charakorn2026doc}.
Building on these advances, MoEGen combines sparse expert-code routing with shared hypernetwork-based LoRA generation, enabling input-dependent specialization while expanding the expert pool without storing a full adapter per expert.

\section{Methodology}
\subsection{Overview}
\label{sec:overview} 

\begin{figure*}[!tbp]
    \centering
    \includegraphics[width=\textwidth]{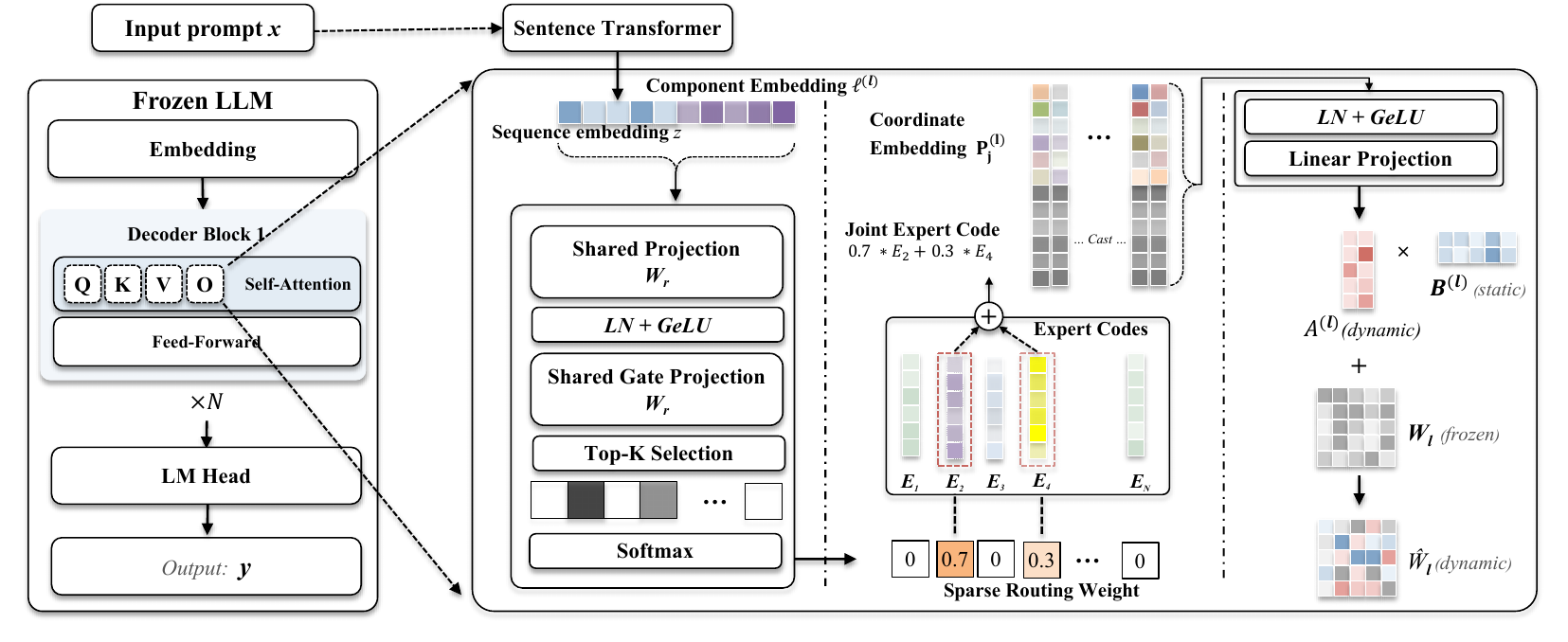}
    \caption{
Overview of MoEGen. A frozen sentence encoder maps the input prompt $x$ to a semantic embedding $z$. For each adapted component $l$, $z$ is concatenated with a learnable component embedding $\ell^{(l)}$ and projected as $h^{(l)} = W_r [z; \ell^{(l)}]$ for sparse expert-code routing. Each selected expert code, together with component-local coordinates $p_j^{(l)}$, conditions a shared hypernetwork to generate a candidate LoRA $A$ matrix. The candidate matrices are then fused using the corresponding routing weights to obtain the input-specific LoRA $A$ matrix.
}

    \label{fig:MoEGen_overview}
\end{figure*}

MoEGen (Fig.~\ref{fig:MoEGen_overview}) consists of four modules: a frozen pretrained LLM, a frozen prompt
encoder, a sparse MoE router, and a shared LoRA hypernetwork. For each input,
the prompt encoder produces a context embedding, which is used by the router to
select a small number of expert codes. These expert codes do not store LoRA
parameters directly. Instead, they serve as compact conditioning vectors that
guide the hypernetwork to generate dynamic LoRA matrices.

We apply MoEGen to the attention projections of each decoder block, including
the query, key, value, and output projections; we refer to each adapted
projection as a \textit{component}. For a component with
frozen weight $W^{(l)} \in \mathbb{R}^{d_{\mathrm{out}} \times d_{\mathrm{in}}}$,
MoEGen generates a low-rank update
$\Delta W^{(l)} \in \mathbb{R}^{d_{\mathrm{out}} \times d_{\mathrm{in}}}$
and computes the adapted output as
\begin{equation}
    y = W^{(l)}x + \frac{\alpha}{r}\Delta W^{(l)}x,
\end{equation}
where $x \in \mathbb{R}^{d_{\mathrm{in}}}$, $y \in \mathbb{R}^{d_{\mathrm{out}}}$,
$r$ is the LoRA rank, and $\alpha$ is the LoRA scaling factor. The low-rank
update is factorized as $\Delta W^{(l)} = B^{(l)} (A^{(l)})^\top$ with
$B^{(l)} \in \mathbb{R}^{d_{\mathrm{out}} \times r}$ and
$A^{(l)} \in \mathbb{R}^{d_{\mathrm{in}} \times r}$, where $r \ll \min(d_{\mathrm{in}}, d_{\mathrm{out}})$. The original LLM parameters are never modified.
\subsection{Per-Component Semantic Routing with Compact Expert Codes}
\label{sec:routing}

MoEGen maintains a single pool of $N$ trainable expert codes
$\{E_1, \ldots, E_N\}$ with $E_i \in \mathbb{R}^{d_h}$, shared across
all $L$ components. The routing parameters remain compact because all components share a
common gating projection $W_g$.

For each input prompt $x$, we use a frozen sentence encoder~\cite{zhang2025qwen3embeddingadvancingtext}
$f_{\mathrm{enc}}: \mathcal{X} \rightarrow \mathbb{R}^{d_z}$ to obtain a
semantic context embedding $z = f_{\mathrm{enc}}(x)$. The encoder remains fixed
during training to preserve general semantic representations and reduce
optimization cost. A lightweight router $f_{\mathrm{r}}$ then maps this shared
semantic embedding into component-specific routing features:

\begin{equation}
\begin{aligned}
f_{\mathrm{r}}(z)
&= [h^{(1)}, \ldots, h^{(L)}],
\hspace{0.5cm}
H =
\begin{bmatrix}
(h^{(1)})^\top & \cdots & (h^{(L)})^\top
\end{bmatrix}^{\top}
\in \mathbb{R}^{L \times d_h}.
\end{aligned}
\end{equation}

where $L$ is the number of adapted LoRA components (i.e., target projections
across decoder layers), and each row $h^{(l)} \in \mathbb{R}^{d_h}$ denotes
the routing feature for component $l$.

To enable component-specific specialization, we associate each component with
a learnable embedding $\ell^{(l)} \in \mathbb{R}^{d_\ell}$ and concatenate it
with the shared prompt embedding:
\begin{equation}
    h^{(l)} = W_r [z; \ell^{(l)}],
\end{equation}
where $[\cdot;\cdot]$ denotes vector concatenation and
$W_r \in \mathbb{R}^{d_h \times (d_z + d_\ell)}$ is a shared projection matrix.
This formulation allows the router to model interactions between prompt-level
semantics and component-specific identities, while requiring only
$d_h(d_z + d_\ell) + L d_\ell$ parameters.

Each expert code $E_i$ is a compact conditioning vector rather than
a full LoRA adapter: it encodes an adaptation style that the
hypernetwork (Sec.~\ref{sec:generation}) later expands into
component-specific LoRA matrices. For each component $l$, the routing feature
$h^{(l)}$ is first passed through a nonlinear block and then projected
to expert logits:
\begin{equation}
\begin{aligned}
    \tilde{h}^{(l)}
    &=
    \mathrm{GELU}\!\left(
        \mathrm{LN}\!\left(h^{(l)}\right)
    \right), 
    g^{(l)}
    &=
    W_g \tilde{h}^{(l)} + b_g.
\end{aligned}
\end{equation}
We then perform top-$k$ selection and normalize the selected logits:
\begin{equation}
\begin{aligned}
\mathcal{T}^{(l)}
&= \operatorname{TopK}\!\left(g^{(l)}, k\right),
w_i^{(l)}
&=
\frac{\exp\!\left(g_i^{(l)}\right)}
{\sum_{j\in\mathcal{T}^{(l)}}
 \exp\!\left(g_j^{(l)}\right)},
\quad i\in\mathcal{T}^{(l)}.
\end{aligned}
\end{equation}
Here, $\mathcal{T}^{(l)} \subseteq \{1,\ldots,N\}$ is the set
of indices corresponding to the $k$ largest entries of $g^{(l)}$.
Experts outside $\mathcal{T}^{(l)}$ are assigned zero weight.

\subsection{Expert-Conditioned LoRA Generation}
\label{sec:generation}

For each adapted component $l$, MoEGen generates an input-specific LoRA
matrix $A^{(l)}$ while maintaining a trainable matrix $B^{(l)}$ shared
across inputs. Motivated by the output-space efficiency considerations in
Text-to-LoRA~\citep{charakorn2025texttolora}, this asymmetric design reduces
generation overhead while retaining input-conditioned adaptation.

Parameter generation is conditioned on compact expert codes and
component-local coordinates. For each component $l$, we introduce a
learnable coordinate embedding
$P^{(l)} \in \mathbb{R}^{d_{\mathrm{in}} \times d_h}$, whose $j$-th row
$p_j^{(l)}$ corresponds to the $j$-th input dimension.
For each selected expert $i \in \mathcal{T}^{(l)}$, a shared hypernetwork
$H_{\phi}$ uses its code $E_i$ and these coordinates to generate a
candidate LoRA matrix $A_i^{(l)}$. The candidate matrices are then fused
using the input-dependent routing weights. Generation and fusion are
defined row-wise as
\begin{equation}
a_{j,i}^{(l)} = H_{\phi}\!\left([p_j^{(l)}; E_i]\right),
\qquad
a_j^{(l)} = \sum_{i \in \mathcal{T}^{(l)}} w_i^{(l)} a_{j,i}^{(l)},
\end{equation}
where $a_{j,i}^{(l)}, a_j^{(l)} \in \mathbb{R}^{r}$ denote the $j$-th
rows of the candidate matrix $A_i^{(l)}$ and the final matrix $A^{(l)}$,
respectively. The resulting input-specific matrix is therefore
$A^{(l)} = \sum_{i \in \mathcal{T}^{(l)}} w_i^{(l)} A_i^{(l)}
\in \mathbb{R}^{d_{\mathrm{in}} \times r}$.

The hypernetwork consists of LayerNorm, GELU activation, and a linear
projection from $2d_h$ to $r$ dimensions. Shared across components and
experts, this block contributes $\mathcal{O}(d_h r)$ parameters,
independent of model depth, $d_{\mathrm{in}}$, $d_{\mathrm{out}}$,
and the number of experts $N$.

Finally, the fused matrix $A^{(l)}$ and the trainable matrix
$B^{(l)} \in \mathbb{R}^{d_{\mathrm{out}} \times r}$ define the low-rank
update $\Delta W^{(l)} = B^{(l)}(A^{(l)})^{\top}$.
Following LoRA~\citep{hu2022lora}, $B^{(l)}$ is initialized to zero
for stable training.

\subsection{Training Objective}
\label{sec:objective}



MoEGen is trained end-to-end with the standard causal language
modeling objective. We optimize three groups of parameters: (i) the
router parameters $W_r$, $\{\ell^{(l)}\}_{l=1}^{L}$, $W_g$, $b_g$;
(ii) the expert codes $\{E_i\}_{i=1}^{N}$; and (iii) the hypernetwork
parameters $\phi_{\mathrm{Linear}}$, $\{P^{(l)}, B^{(l)}\}_{l=1}^{L}$. The
base model weights remain frozen throughout training. To avoid routing collapse and encourage balanced expert utilization, we add a load-balancing regularization loss following~\cite{fedus2022switchtransformersscalingtrillion}.
The full training objective is

\begin{equation}
\mathcal{L}
=
\mathcal{L}_{\mathrm{LM}}
+
\lambda_{\mathrm{lb}}\mathcal{L}_{\mathrm{load}},
\end{equation}

where the language modeling loss is
$\mathcal{L}_{\mathrm{LM}}
=
-\sum_{t=1}^{T}
\log p_{\theta}(y_t \mid y_{<t}, X).
$
given an input-output pair $(X,Y)$,
$\lambda_{\mathrm{lb}}$ controls the strength of the load-balancing
regularization. The load-balancing loss is defined as

\begin{equation}
\begin{aligned}
\mathcal{L}_{\mathrm{load}}
=
\frac{N}{L}
\sum_{l=1}^{L}\sum_{i=1}^{N}
&\left(
\frac{1}{Bk}\sum_{b=1}^{B}
\mathbf{1}[i\in\mathcal{T}_b^{(l)}]
\right) 
&\times
\left(
\frac{1}{B}\sum_{b=1}^{B}
w_{b,i}^{(l)}
\right).
\end{aligned}
\end{equation}

where $\mathcal{T}_b^{(l)}$ denotes the top-$k$ expert set selected for the
$b$-th input at component $l$, and $w_{b,i}^{(l)}$ is the normalized routing weight
for expert $i$, with $w_{b,i}^{(l)}=0$ if $i\notin\mathcal{T}_b^{(l)}$. This regularizer couples the discrete routing decisions with
the differentiable routing probabilities, encouraging more balanced expert
usage across both inputs and components.

\definecolor{bestrow}{RGB}{255, 237, 204}


  \section{Experiment}

\begin{table*}[!bp]
\centering
\caption{
Performance comparison on eight commonsense reasoning benchmarks.
All benchmark results are accuracy scores (\%).
Params (\%) denotes the percentage of trainable parameters.
Average denotes the macro-average across the eight benchmarks.
The best score for each benchmark and the highest average within
each base model are highlighted in bold.
}
\label{tab:main_results}

\begingroup
\small
\setlength{\tabcolsep}{3.5pt}
\renewcommand{\arraystretch}{0.98}
\setlength{\aboverulesep}{1.5pt}
\setlength{\belowrulesep}{1.5pt}
\setlength{\heavyrulewidth}{0.6pt}
\setlength{\lightrulewidth}{0.4pt}
\setlength{\cmidrulewidth}{0.4pt}
\setlength{\arrayrulewidth}{0.4pt}

\begin{adjustbox}{max width=\textwidth}
\begin{tabular}{@{}c|c|c|*{9}{c}@{}}
\toprule
\textsc{Model}
& \textsc{Method}
& Params (\%)
& BoolQ
& PIQA
& SIQA
& HellaS.
& WinoG.
& ARC-e
& ARC-c
& OBQA
& \textbf{Average} \\
\midrule

\multirow{7}{*}{LLaMA2-7B}
 & LoRA
 & 0.83 & 69.8 & 79.9 & 79.5 & 83.6
 & 82.6 & 79.8 & 64.7 & 81.0 & 77.6 \\

 & DoRA
 & 0.42 & 72.0 & 83.1 & 79.9 & 89.1
 & 83.0 & 84.5 & 71.0 & 81.2 & 80.5 \\

 & SpIEL
 & 0.83 & 70.5 & 80.6 & 80.8 & 85.8
 & 83.4 & 81.2 & 65.8 & 81.8 & 78.7 \\

 & SMT
 & 0.84 & 72.0 & 83.8 & 80.8 & 93.3
 & 82.8 & 86.7 & 74.0 & 81.0 & 81.8 \\

 & MoELoRA
 & 1.19 & \sota{73.8} & 85.5 & \sota{81.3}
 & \sota{94.8} & \sota{85.9}
 & 87.9 & 75.0 & 84.2 & 83.6 \\

 & MoLA
 & 0.87 & 71.8 & 81.9 & 78.6 & 88.9
 & 81.6 & 82.8 & 69.5 & 80.0 & 79.4 \\

\cmidrule{2-12}
 & \textsc{MoEGen} (Ours)
 & 0.75 & 73.3 & \sota{86.4} & \sota{81.3}
 & \sota{94.8} & \sota{85.9} & \sota{88.9}
 & \sota{75.4} & \sota{87.4} & \sota{84.2} \\
\midrule

\multirow{7}{*}{LLaMA3-8B}
 & LoRA
 & 0.70 & 70.8 & 85.2 & 79.9 & 91.7
 & 84.3 & 84.2 & 71.2 & 79.0 & 80.8 \\

 & DoRA
 & 0.71 & 74.6 & 89.3 & 79.9 & 95.5
 & 85.6 & 90.5 & 80.4 & 85.8 & 85.2 \\

 & SpIEL
 & 0.70 & 72.1 & 83.6 & 80.0 & 91.8
 & 85.4 & 91.2 & 76.8 & 80.8 & 82.7 \\

 & SMT
 & 0.71 & 75.7 & 88.4 & 81.4 & 96.2
 & 88.2 & 92.7 & 83.2 & 88.6 & 86.8 \\

 & MoELoRA
 & 1.05 & 74.7 & 88.9 & 81.3 & 96.4
 & \sota{89.3} & 92.3 & 81.4
 & \sota{89.6} & 86.7 \\

 & MoLA
 & 0.73 & 73.9 & 86.8 & 79.2 & 93.9
 & 86.0 & 87.6 & 75.9 & 83.2 & 83.3 \\

\cmidrule{2-12}
 & \textsc{MoEGen} (Ours)
 & 0.47 & \sota{76.1} & \sota{90.9} & \sota{83.0}
 & \sota{96.6} & 89.1 & \sota{93.3}
 & \sota{84.4} & 89.4 & \sota{87.9} \\
\midrule

\multirow{7}{*}{Qwen3-8B-base}
 & LoRA
 & 0.70 & 74.3 & 89.0 & 80.3 & 94.4
 & 88.0 & 94.7 & 88.1 & 90.4 & 87.4 \\

 & DoRA
 & 0.71 & 72.9 & 90.9 & 82.7 & 95.9
 & 88.9 & 97.0 & 90.9 & \sota{93.2} & 89.1 \\

 & SpIEL
 & 1.00 & 72.8 & 88.2 & 79.8 & 93.8
 & 88.2 & 95.8 & 87.2 & 87.8 & 86.7 \\

 & SMT
 & 0.71 & 74.9 & 88.9 & 81.9 & 94.8
 & 86.6 & 95.5 & 88.4 & 92.2 & 87.9 \\

 & MoELoRA
 & 0.75 & 72.5 & 90.6 & 81.9 & 93.4
 & 84.1 & \sota{97.9} & \sota{93.6}
 & 89.8 & 88.0 \\

 & MoLA
 & 1.50 & 74.3 & 90.4 & 80.9 & 95.3
 & 86.7 & 96.8 & 91.9 & 92.2 & 88.6 \\

\cmidrule{2-12}
 & \textsc{MoEGen} (Ours)
 & 0.52 & \sota{75.6} & \sota{91.4} & \sota{83.0}
 & \sota{96.4} & \sota{90.0} & 97.7
 & 92.4 & 92.6 & \sota{89.9} \\

\bottomrule
\end{tabular}
\end{adjustbox}
\endgroup
\end{table*}

\label{sec:experiment}
\subsection{Experiment Setup} 
\noindent\textbf{Datasets and benchmarks.} Following~\cite{he2025smt}, we use
\textsc{Commonsense-170K}, which combines the training splits of eight
commonsense reasoning benchmarks: BoolQ~\cite{clark2019boolqexploringsurprisingdifficulty}, PIQA~\cite{bisk2019piqareasoningphysicalcommonsense}, SIQA~\cite{sap-etal-2019-social}, HellaSwag~\cite{zellers2019hellaswagmachinereallyfinish}, WinoGrande~\cite{sakaguchi2019winograndeadversarialwinogradschema},
ARC-Easy~\cite{allenai:arc}, ARC-Challenge~\cite{allenai:arc}, and OpenBookQA~\cite{mihaylov-etal-2018-suit}. We evaluate each method on the
held-out test set of each benchmark. We also use a mixed-task dataset composed of four biomedical and legal-domain
tasks: MedNLI~\cite{romanov-shivade-2018-lessons}, PubMedQA~\cite{jin2019pubmedqadatasetbiomedicalresearch}, HQS~\cite{ben-abacha-demner-fushman-2019-summarization}, and BillSum~\cite{kornilova-eidelman-2019-billsum}. MedNLI and PubMedQA are evaluated
by accuracy, while HQS and BillSum are evaluated by ROUGE-1, ROUGE-2, and
ROUGE-L ~\cite{lin-2004-rouge}. For this setting, the \textit{Avg} column averages one primary metric
per task: accuracy for MedNLI and PubMedQA, and ROUGE-L for HQS and BillSum.

\noindent\textbf{Baselines.}
  We compare MoEGen with static and mixture-of-experts PEFT
  methods. The static baselines include LoRA, DoRA, SpIEL, and SMT,
  whose adaptation parameters remain fixed across inputs
  at inference time. The MoE-based baselines include MoELoRA
  and MoLA, which incorporate expert routing into low-rank
  adaptation. These comparisons assess the effectiveness
  of input-conditioned parameter generation relative to
  static adaptation and existing expert-based approaches.

\noindent\textbf{Implementation details.} MoEGen uses a frozen Qwen3-Embedding-0.6B~\cite{zhang2025qwen3embeddingadvancingtext} sentence encoder.
We use $N{=}8$ expert codes with code dimension $d_h{=}32$, top-$k{=}2$ routing, and LoRA rank $r{=}64$, and apply the generated adapters to the Q/K/V/O projections
of every decoder block. All methods are trained for $3$ epochs with AdamW~\cite{loshchilov2017decoupled} and
bfloat16 precision. The effective batch size is $16$ for commonsense reasoning
and $64$ for mixed-task joint training. Additional details are provided in Appendix.

\subsection{Commonsense Reasoning}
\label{sec:main_results}

Table~\ref{tab:main_results} reports the results across
LLaMA-2-7B~\cite{touvron2023llama2openfoundation},
LLaMA-3-8B~\cite{grattafiori2024llama3herdmodels}, and
Qwen3-8B-base~\cite{yang2025qwen3technicalreport}. MoEGen achieves
the highest average score on all three backbones, reaching $84.2$, $87.9$,
and $89.9$, respectively. Compared with the strongest static PEFT baseline
on each backbone, MoEGen improves the average score by $2.4$ points
over SMT on LLaMA-2-7B, by $1.1$ points over SMT on LLaMA-3-8B, and by
$0.8$ points over DoRA on Qwen3-8B-base. 
MoEGen trains only $0.75\%$, $0.47\%$, and $0.52\%$ of the backbone parameters on LLaMA2-7B, LLaMA3-8B,
  and Qwen3-8B-Base, respectively. These parameter budgets are smaller than those of both MoLA and
  MoELoRA, indicating that the improvements are not due to a larger trainable parameter budget.

Compared with standard LoRA, MoEGen achieves a higher score on
every evaluated benchmark across all three backbones. On LLaMA-2-7B, it
improves the average score from $77.6$ to $84.2$. On LLaMA-3-8B, the
average score increases from $80.8$ to $87.9$. On Qwen3-8B-base, where
LoRA achieves an average score of $87.4$, MoEGen reaches $89.9$.
These results show that input-conditioned adaptation provides consistent
benefits across model families with different baseline accuracies. Notably,
the improvement remains positive even on Qwen3-8B-Base, where standard
LoRA already performs strongly, suggesting that MoEGen captures useful
input-dependent specialization beyond a single static low-rank update.

MoEGen also achieves higher average scores than the MoE-based PEFT
methods MoELoRA and MoLA. MoELoRA and MoLA increase adapter capacity by
introducing multiple LoRA experts, with each expert corresponding to a
fixed set of learned parameters. In contrast, MoEGen routes among
compact expert codes and uses the resulting conditioning signal to
synthesize input-specific LoRA matrices. MoEGen improves over
MoELoRA by $0.6$, $1.2$, and $1.9$ points on LLaMA-2-7B, LLaMA-3-8B,
and Qwen3-8B-base, respectively. Compared with MoLA, the corresponding
improvements are $4.8$, $4.6$, and $1.3$ points. The results indicate
that generating input-specific adapter parameters yields higher average
accuracy than selecting or mixing a fixed set of adapter experts in
these experiments.

\begin{figure*}[!tbp]
    \centering
    \includegraphics[width=\textwidth]
      {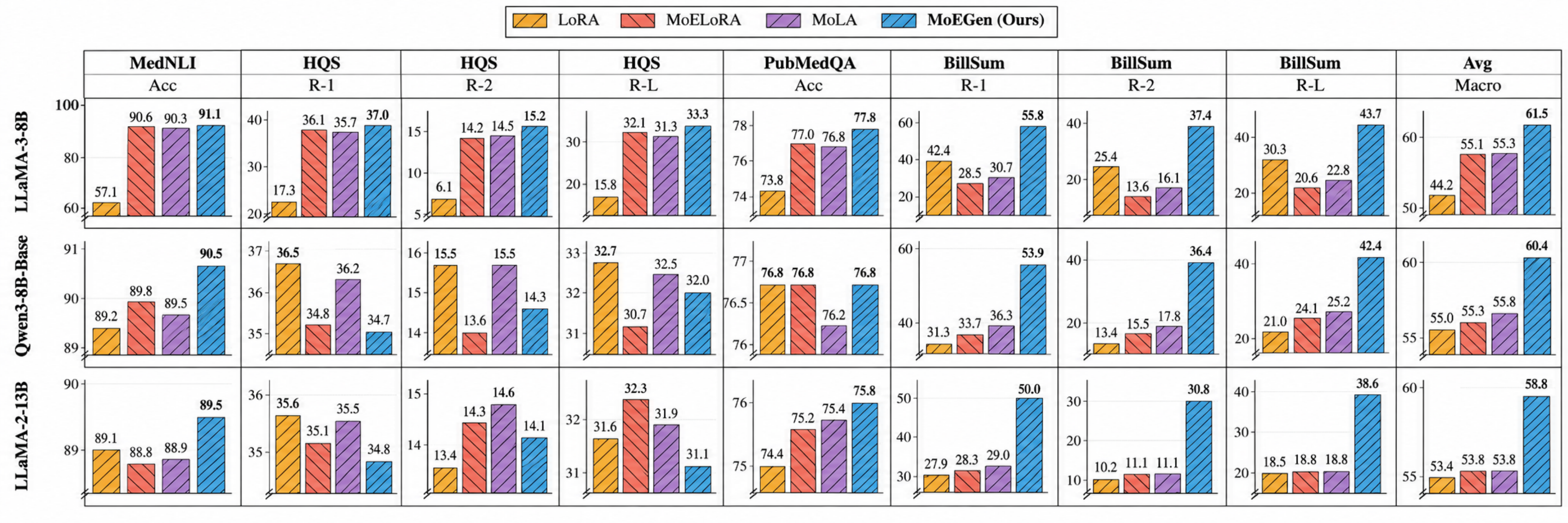}
    \caption{
      Comparison on the cross-domain benchmark at epoch 3.
      Rows correspond to LLaMA-3-8B, Qwen3-8B-Base, and LLaMA-2-13B.
      MedNLI and PubMedQA report accuracy; HQS and BillSum report
      ROUGE-1, ROUGE-2, and ROUGE-L.
      Avg is the macro-average of the two accuracy scores and
      the two ROUGE-L scores. Bold values indicate the
      best results within each panel.
      Additional numerical results are provided in the appendix. }
    \label{fig:nlp_joint_comparison}
    \vspace{-5mm}
  \end{figure*}

\subsection{Cross-Domain Joint Training}
\label{sec:nlp_joint}

To evaluate MoEGen under heterogeneous multi-task adaptation, we
jointly train on MedNLI, PubMedQA, HQS, and BillSum following
\cite{lu2024allinonetuningstructuralpruning}. These tasks cover natural
language inference, question answering, and clinical and legal summarization.
All methods use the same training data and optimization protocol and are
evaluated separately on each task. Figure~\ref{fig:nlp_joint_comparison} reports
the results.

On LLaMA-3-8B, MoEGen achieves the highest score on every metric
and improves the average score by $6.2$ points over MoLA, from $55.3\%$
to $61.5\%$. On BillSum, MoEGen reaches $43.7\%$ ROUGE-L,
compared with $30.3\%$ for the best baseline. Plain LoRA performs worst:
it drops to $57.1\%$ on MedNLI and $15.8\%$ on HQS, which indicates
that a single static update struggles to accommodate the competing
adaptation requirements of all four tasks. The advantage holds on Qwen3-8B-Base. MoEGen reaches the highest
average of $60.4\%$, compared with $55.8\%$ for the strongest baseline
(MoLA), obtains the best MedNLI accuracy, and matches the best PubMedQA
accuracy. On BillSum it reaches $42.4\%$ ROUGE-L against $25.2\%$ for the
best baseline, while on HQS it stays within $0.7$ points of the best
result. 

MoEGen also achieves the highest average score on LLaMA-2-13B,
reaching $58.8\%$ compared with $53.8\%$ for MoELoRA and MoLA. It
obtains the highest MedNLI and PubMedQA accuracy and improves BillSum
ROUGE-L from $18.8\%$ to $38.6\%$. On HQS, its scores remain within
$1.2$ points of the best baseline across all three ROUGE metrics.
Together, the results on all three backbones show that MoEGen
consistently improves overall performance under cross-domain joint
training.

\subsection{Ablation Study}
\label{sec:ablation_num_codes}

\begin{figure}[t]
    \centering
    \includegraphics[width=\linewidth]
        {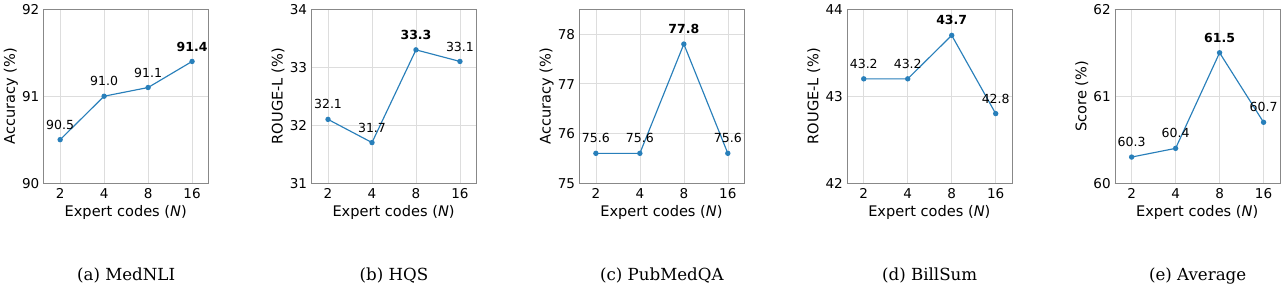}
    \caption{
        Effect of the expert-code pool size on the
        cross-domain joint benchmark with LLaMA-3-8B.
        Panels report accuracy for MedNLI and PubMedQA,
        ROUGE-L for HQS and BillSum, and the average
        score across the four tasks.
        All scores are shown on a percentage scale,
        with the best value in each panel bolded.
        Each panel uses its own vertical-axis range.
    }
    \label{fig:ablation_num_codes}
    \vspace{-5mm}
\end{figure}

\paragraph{Number of Expert.} We study the effect of the expert-code pool size
$N \in \{2,4,8,16\}$ on the cross-domain joint benchmark using
LLaMA-3-8B. All variants follow the setup in
Section~\ref{sec:nlp_joint}, with $r{=}64$, $d_h{=}32$, top-$2$
routing, an effective batch size of $64$, and $3$ training epochs.
As shown in Figure~\ref{fig:ablation_num_codes}, $N{=}8$ achieves the
highest average score and performs best on three of the four tasks.
Smaller pools may provide insufficient capacity for heterogeneous domains,
whereas a larger pool does not yield further improvements. Notably, when
$N{=}2$ under top-$2$ routing, both codes are always selected, reducing the
router to weighting a fixed code pair. Its lower performance suggests that
input-dependent code selection contributes to the effectiveness of
MoEGen. We therefore use $N{=}8$ for the cross-domain benchmark. An ablation study on the routing design and shared expert is also provided in Appendix~\ref{sec:ablation_topk_shared}.

\begin{figure*}[!tbp]
    \centering

    \begin{minipage}[c]{0.40\textwidth}
        \centering
        \includegraphics[width=\linewidth]
        {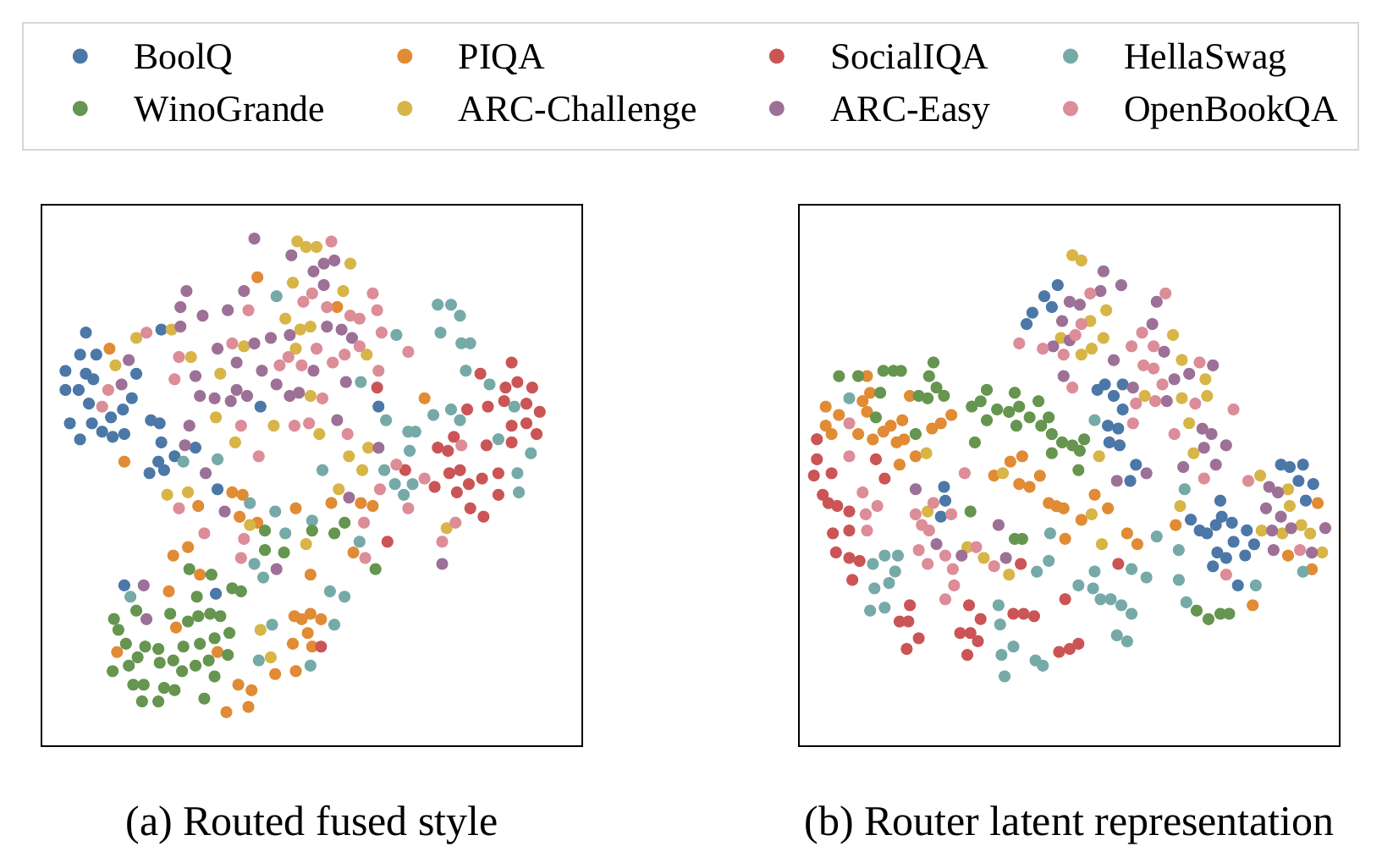}
        \caption{
        Case study of MoEGen routing on eight commonsense reasoning test sets.
        \textbf{(a)} t-SNE of the routed fused style produced by the hypernetwork.
        \textbf{(b)} t-SNE of the router latent representation used by the gate.
        }
        \label{fig:case-study}
        \vspace{-5mm}
    \end{minipage}
    \hfill
    \begin{minipage}[c]{0.57\textwidth}
        \centering
        \setlength{\tabcolsep}{2pt}
        \setlength{\aboverulesep}{1pt}
        \setlength{\belowrulesep}{1pt}
        \renewcommand{\arraystretch}{0.9}

        \refstepcounter{table}
        \label{tab:ablation_ctx_encoder}

        \captionof*{table}{
        \textbf{Table \thetable(a):}
        Ablation of the context encoder and MoE routing using LLaMA3-8B
        as the base model. Without the frozen encoder, routing uses the
        backbone's last-token representation. Without MoE routing,
        the model uses a single hypernetwork.
        }

        \scriptsize
        \begin{tabular*}{\linewidth}{@{\extracolsep{\fill}}lcccc|c@{}}
        \toprule
        Method & MedNLI & BillSum & HQS & PubMedQA & Avg \\
               & Acc    & R-L     & R-L & Acc      &     \\
        \midrule
        MoEGen
        & \textbf{91.1\%}
        & \textbf{43.7\%}
        & \textbf{33.3\%}
        & \textbf{77.8\%}
        & \textbf{61.5\%} \\
        \;\;w/o sentence encoder
        & 90.8\%
        & 43.6\%
        & 32.4\%
        & 76.4\%
        & 60.8\% \\
        \;\;w/o MoE
        & 90.7\%
        & 40.5\%
        & 30.9\%
        & 77.0\%
        & 59.8\% \\
        \bottomrule
        \end{tabular*}

        \vspace{2mm}

        \phantomsection
        \label{tab:encoder_efficiency}

\vspace{-2mm}

\phantomsection
\label{tab:encoder_efficiency}

\captionof*{table}{
\textbf{Table \thetable(b):}
Comparison of the context encoder with respect to training time, inference latency, and memory usage. Bold indicates lower time, latency, or memory usage.}

\scriptsize
\begin{tabular*}{\linewidth}{
@{\extracolsep{\fill}}lcccc@{}
}
\toprule
Method
& Train Time $\downarrow$
& Latency $\downarrow$
& Memory $\uparrow$
& Frozen Params. \\
& (min)
& (ms/ex.)
& (GB)
& \\
\midrule

MoEGen
& \textbf{13.1}
& \textbf{24.6}
& 19.3
& 0.6B \\

\;\;w/o sentence encoder
& 14.2
& 25.2
& \textbf{17.1}
& \textbf{N/A} \\

\bottomrule
\end{tabular*}

    \end{minipage}

    \vspace{-5mm}
\end{figure*}

\paragraph{Context encoding and MoE routing.}
  We study the contributions of the context encoder and MoE
  routing
  in Table~\ref{tab:ablation_ctx_encoder} (a). By default, MoEGen
  uses the
  frozen \texttt{Qwen3-Embedding-0.6B} encoder to obtain the
  routing
  context \(z\). Replacing it with the final-token
  representation
  of the frozen 8B backbone (\texttt{w/o sentence encoder})
  reduces
  the average score from 61.5\% to 60.8\%, with the largest
  drops on
  PubMedQA (77.8\% to 76.4\%) and HQS (33.3\% to 32.4\%). This
  suggests
  that the external encoder provides more effective routing
  context
  under the evaluated configuration. 
 We also remove the MoE routing module and retain only the input-conditioned HyperNetwork (w/o MoE). The average score drops to 59.8\%, with larger decreases on BillSum (43.7\% to 40.5\%) and HQS (33.3\% to 30.9\%). This suggests that expert routing provides additional gains beyond input-conditioned HyperNetwork generation alone.

  As shown in Table \ref{tab:ablation_ctx_encoder}(b), the external encoder also offers a favorable time--memory
  trade-off.
  Compared with obtaining context from the 8B backbone, it
  reduces
  training time from 14.2 to 13.1 minutes (approximately 8\%)
  and
  inference latency from 25.2 to 24.6\,ms per example by
  avoiding an
  additional forward pass through the larger backbone. This
  requires
  0.6B additional frozen parameters and increases GPU memory
  usage
  from 17.1 to 19.3\,GB. We therefore retain the external
  encoder
  and MoE routing in our default configuration.

\subsection{Qualitative Study}

We visualize the routing behavior of MoEGen on eight commonsense
reasoning benchmarks. For each task, we randomly sample 50 test prompts and
apply t-SNE to the router inputs and the fused expert representations.

As shown in Figure~\ref{fig:case-study}, the router inputs already exhibit
task-dependent structure. Distinctive tasks such as SocialIQA, WinoGrande,
and BoolQ form relatively compact regions, whereas science QA benchmarks
such as ARC-Challenge, ARC-Easy, and OpenBookQA are more closely distributed. After expert mixing, the fused representations preserve this organization
while connecting related tasks more smoothly. This suggests that the router
captures both task-specific adaptation patterns and shared structures across
similar reasoning tasks. Further experimental analyses, including additional qualitative examples
and an evaluation of computational efficiency, are provided in
Appendix~\ref{qual_study_additional}.


\section{Conclusion}

This paper proposes an instance-adaptive framework for
parameter-efficient LLM adaptation. It aligns LoRA updates with input semantics
through MoE-guided routing and dynamic parameter generation driven by a
lightweight hypernetwork. By replacing full LoRA experts with compact expert
codes, MoEGen expands the adaptation space while keeping the parameter
cost low. Experimental results show that MoEGen consistently outperforms
static LoRA variants and existing MoE-based PEFT methods across commonsense,
biomedical, and legal-domain benchmarks, demonstrating its advantages in
accuracy and adaptation flexibility.

\clearpage

\subsection*{AI use statement}


In this work, we used generative AI tools for language polishing. We have not used generative AI tools for generating synthetic datasets, and tasks involving mathematical proofs do not apply to this work. We have reviewed all AI-assisted outputs. We take responsibility for the final content of this work, including text, claims or artifacts produced with the aid of generative AI.

\subsection*{Ethics statement}

Our training data include existing biomedical datasets, namely MedNLI, PubMedQA, and HQS, which have been released for research use. We acknowledge the sensitive nature of medical applications and the potential risks of inaccurate model outputs. Our experiments evaluate model adaptation on these benchmarks and do not establish suitability for clinical use.

\subsection*{Reproducibility statement}

To facilitate reproducibility, we describe the proposed method in Section 3 and provide the datasets, evaluation metrics, and experimental settings in Section 4.1. Appendix A.1 further specifies the implementation details and baseline configurations. Unless otherwise specified, all methods are trained for 3 epochs using AdamW with a learning rate of \(2 \times 10^{-4}\), cosine learning-rate decay, and the Alpaca prompt format, with the backbone frozen and only adapter-related parameters updated. We will publicly release our code upon acceptance.

\subsubsection*{Author Contributions}

\noindent\textbf{Project Leader:}
Yiming Zeng$^{1,\dagger}$.

\noindent\textbf{Contributors:}
Yiming Zeng$^{1,\dagger}$,
Lei Lu$^{2,\dagger}$,
Zexin Li$^{3,*}$.
Zhuochun Li$^{4}$,
Dehai Min$^{5}$,
Shuoqiu Li$^{6}$,
Shuyi Liao$^{1}$,
Xidong Wu$^{7}$,
Zeyu Zhang$^{8}$,
Minmei Wang$^{1}$,
Yu Zhao$^{9}$,
Tingting Yu$^{1,*}$,
Shangqian Gao$^{6,*}$.

\noindent\textbf{Corresponding Authors:}
Zexin Li$^{3,*}$,
Tingting Yu$^{1,*}$,
Shangqian Gao$^{6,*}$.

\noindent\textbf{Author Affiliations:}
$^{1}$University of Connecticut \quad
$^{2}$Northeastern University \quad
$^{3}$Nanyang Technological University \quad
$^{4}$University of Pittsburgh \quad
$^{5}$University of Illinois at Chicago \quad
$^{6}$Florida State University \quad
$^{7}$Google \quad
$^{8}$Amazon \quad
$^{9}$University of Cincinnati


\bibliography{iclr2027_conference}
\bibliographystyle{iclr2027_conference}

\appendix
\section{Appendix}
\subsection{Implementation Details}

\label{app:implementation}

We keep the backbone LLM frozen for all PEFT methods and update only the
adapter-related parameters. Unless otherwise specified, all methods are trained
for 3 epochs with AdamW~\cite{loshchilov2017decoupled}, a learning rate of
$2\times 10^{-4}$, cosine learning-rate decay, and the Alpaca
prompt format.

For static PEFT baselines, including LoRA, DoRA, and SMT, we follow the
experimental setup of SMT for training-related hyperparameters, including the
learning rate, number of epochs, warmup steps, scheduler, and prompt format. We
keep each method's adaptation modules consistent with the SMT setting whenever
applicable: LoRA and DoRA are applied to the Q/K/V/O attention projections and
the Gate/Up/Down feed-forward projections, while SMT is applied to the Q/K/V
attention projections.

For the MoE-LoRA and MoLA baselines, we generally follow the hyperparameter
settings used in their original papers and keep the implementation consistent
with their routing designs. Specifically, MoE-LoRA uses $N{=}4$ LoRA experts
with rank $r{=}32$ and a task-conditioned soft mixture, while MoLA uses
$N{=}8$ LoRA experts with rank $r{=}32$ and token-level top-$k$ routing with
$k{=}2$. Both baselines are applied to the Q/K/V/O attention projections and
the Gate/Up/Down feed-forward projections.

\subsection{Ablation Study on Routing Design Choices} 
\label{sec:ablation_topk_shared}

\begin{table*}[!tbp]
  \centering
  \caption{Additional BillSum qualitative examples illustrating two distinct
  failure modes of the static LoRA-based baselines: a repetitive-loop failure
  on the Comprehensive Immunosuppressive Drug Coverage Act (top) and a
  mid-clause-boilerplate failure on the Federal Election Law Reform
  Commission Act (bottom). All methods are trained jointly on the
  cross-domain mixture (MedNLI, PubMedQA, HQS, BillSum) with Llama-2-13B.}
  \label{tab:qual_app_billsum}
  \renewcommand{\arraystretch}{1.3}
  \small
  \begin{tabular}{p{0.16\textwidth} c p{0.66\textwidth}}
  \toprule
  \textbf{Method} & \textbf{R-L} & \textbf{Predicted summary} \\
  \midrule
  \multicolumn{3}{l}{\textit{\textbf{Example 1: Comprehensive Immunosuppressive Drug Coverage Act (repetitive-loop failure).}}} \\
  \midrule
  Reference & --
  & Comprehensive Immunosuppressive Drug Coverage for Kidney Transplant
    Patients Act of 2016. This bill amends titles II and XVIII (Medicare) of
    the Social Security Act to indefinitely extend Medicare coverage of
    immunosuppressive drugs for kidney transplant recipients. Under current
    law, such coverage is limited to 36 months following a transplant. \\
  LoRA & 0.182
  & \textit{(c) of this section shall apply to such an individual in the same
    manner as they apply to an individual who is entitled to benefits under
    part A.} \\
  MoE-LoRA & 0.047
  & \textit{(c) of this section shall not apply to such an individual. The
    provisions of subsection (d) of this section shall not apply to such an
    individual\ldots The provisions of subsection (e) of this section shall
    not apply to such an individual\ldots} (the pattern repeats for
    subsections (f), (g), \ldots) \\
  MoLA & 0.069
  & \textit{(c) shall apply to such individual's coverage period in the same
    manner as they apply to the coverage period of an individual who is
    entitled to benefits under part A. The provisions of subsection (d)
    shall apply to such individual's coverage period\ldots The provisions of
    subsection (e) shall apply\ldots} (the pattern repeats for subsections
    (f), (g), \ldots) \\
  \textbf{MoEGen (Ours)} & \textbf{0.571}
  & Comprehensive Immunosuppressive Drug Coverage for Kidney Transplant
    Patients Act of 2016. This bill amends title XVIII (Medicare) of the
    Social Security Act to require the Medicare program to provide
    immunosuppressive drug coverage for kidney transplant recipients. The
    bill also requires the program to provide immunosuppressive drug coverage
    for individuals whose entitlement to Medicare Part A hospital insurance
    benefits has ended by reason of the 36-month limit on such benefits. \\
  \midrule
  \multicolumn{3}{l}{\textit{\textbf{Example 2: Federal Election Law Reform Commission Act (mid-clause-boilerplate failure).}}} \\
  \midrule
  Reference & --
  & Federal Election Law Reform Commission Act - Establishes the Federal
    Election Law Reform Commission to recommend Federal election law reform.
    Requires a Federal election bill to be introduced in the House and the
    Senate within a specified time of the Commission's submittal of draft
    legislation. Authorizes appropriations. \\
  LoRA & 0.135
  & \textit{\ldots be considered as part of the rules of each House,
    respectively, and such rules shall supersede other rules only to the
    extent that they are inconsistent therewith.} \\
  MoE-LoRA & 0.135
  & \textit{\ldots be considered as part of the rules of each House,
    respectively, and such rules shall supersede other rules only to the
    extent that they are inconsistent therewith.} \\
  MoLA & 0.139
  & \textit{\ldots be considered a part of the rules of each House,
    respectively, and shall supersede other rules only to the extent that
    they are inconsistent therewith.} \\
  \textbf{MoEGen (Ours)} & \textbf{0.522}
  & Federal Election Law Reform Commission Act - Establishes the Federal
    Election Law Reform Commission to study and recommend reforms in Federal
    election laws. \\
  \bottomrule
  \end{tabular}
\end{table*}

\begin{table*}[!tbp]
  \centering
  \caption{Qualitative example on BillSum (\textit{Heavy Truck Tax Fairness Act
  of 2010}).}
  \label{tab:qual_billsum}
  \renewcommand{\arraystretch}{1.08}
  \small
  \setlength{\tabcolsep}{4pt}
  \begin{tabular}{p{0.18\textwidth} p{0.76\textwidth}}
  \toprule
  \textbf{Method} & \textbf{Predicted summary} \\
  \midrule
  Reference
  & Heavy Truck Tax Fairness Act of 2010 - Amends the Internal Revenue Code to:
    (1) repeal the excise tax on heavy trucks and trailers sold at retail; (2)
    increase the rate of tax on diesel fuel or kerosene to 31.6 cents per
    gallon; and (3) impose a floor stocks tax with respect to certain taxable
    fuels. \\
\midrule
  LoRA
  & \textit{und of such tax is not allowable under section 6427(e) of the
    Internal Revenue Code of 1986.} \\
\midrule
  MoE-LoRA
  & \textit{und of such tax is allowable under section 6427 of the Internal
    Revenue Code of 1986.} \\
\midrule
  MoLA
  & \textit{und of such tax is allowable under section 6427 of the Internal
    Revenue Code of 1986.} \\
\midrule
  \textbf{MoEGen (Ours)}
  & Heavy Truck Tax Fairness Act of 2010 - Amends the Internal Revenue Code to
    repeal the excise tax on heavy trucks and trailers. Increases the tax on
    diesel fuel and kerosene. Imposes a floor stocks tax on any taxable liquid
    held on the enactment date by any person. \\
  \bottomrule
  \end{tabular}
  \vspace{-5mm}
\end{table*}

We ablate two routing design choices in MoEGen: the number of activated
experts and the use of an always-on shared expert. All variants are evaluated on
the cross-domain joint benchmark with LLaMA-2-13B as the backbone, following the
same training setup as Section~\ref{sec:nlp_joint}. They use LoRA rank
$r{=}64$, scaling factor $\alpha{=}128$, expert dimension $32$, and $3$ epochs of supervised fine-tuning. The variants differ only in the
routing strategy.


Table~\ref{tab:ablation_topk_shared} shows that top-2 routing achieves the best
overall performance, improving the average score from $0.5844$ to $0.5877$
over top-1 routing. The gains on MedNLI, BillSum, and PubMedQA suggest that
combining two routed expert codes provides a more flexible adaptation space than
selecting a single expert. HQS is the only exception, where top-1 routing gives
a slightly higher ROUGE-L score, possibly because its summarization format is
more constrained and benefits from stronger routing sparsity.

 \begin{table*}[!tbp]
    \centering
    \caption{Comparison on the cross-domain benchmark with
      LLaMA-3-8B, Qwen3-8B-Base, and LLaMA-2-13B at epoch 3.
      All scores are percentages, and best results for each model
      are in \textbf{bold}.
      Avg is the unweighted mean of MedNLI/PubMedQA accuracy
      and HQS/BillSum ROUGE-L.}
    \label{tab:nlp_joint_full}

    \scriptsize
    \setlength{\tabcolsep}{3.2pt}
    \renewcommand{\arraystretch}{0.95}
    \resizebox{0.98\textwidth}{!}{
      \begin{tabular}{l l c c c c c c c c c}
        \toprule
        \multirow{2}{*}{Model} & \multirow{2}{*}{Method}
        & MedNLI
        & \multicolumn{3}{c}{HQS}
        & PubMedQA
        & \multicolumn{3}{c}{BillSum}
        & \multirow{2}{*}{Avg} \\
        \cmidrule(lr){3-3}
        \cmidrule(lr){4-6}
        \cmidrule(lr){7-7}
        \cmidrule(lr){8-10}
        & & Acc
        & R-1 & R-2 & R-L
        & Acc
        & R-1 & R-2 & R-L
        & \\
        \midrule

        \multirow{4}{*}{LLaMA-3-8B}
        & LoRA
        & 57.1\%
        & 17.3\% & 6.1\% & 15.8\%
        & 73.8\%
        & 42.4\% & 25.4\% & 30.3\%
        & 44.2\% \\
        & MoELoRA
        & 90.6\%
        & 36.1\% & 14.2\% & 32.1\%
        & 77.0\%
        & 28.5\% & 13.6\% & 20.6\%
        & 55.1\% \\
        & MoLA
        & 90.3\%
        & 35.7\% & 14.5\% & 31.3\%
        & 76.8\%
        & 30.7\% & 16.1\% & 22.8\%
        & 55.3\% \\
        & \textbf{MoEGen}
        & \textbf{91.1\%}
        & \textbf{37.0\%} & \textbf{15.2\%} & \textbf{33.3\%}
        & \textbf{77.8\%}
        & \textbf{55.8\%} & \textbf{37.4\%} & \textbf{43.7\%}
        & \textbf{61.5\%} \\
        \midrule

        \multirow{4}{*}{Qwen3-8B-Base}
        & LoRA
        & 89.2\%
        & \textbf{36.5\%} & \textbf{15.5\%} & \textbf{32.7\%}
        & \textbf{76.8\%}
        & 31.3\% & 13.4\% & 21.0\%
        & 55.0\% \\
        & MoELoRA
        & 89.8\%
        & 34.8\% & 13.6\% & 30.7\%
        & \textbf{76.8\%}
        & 33.7\% & 15.5\% & 24.1\%
        & 55.3\% \\
        & MoLA
        & 89.5\%
        & 36.2\% & \textbf{15.5\%} & 32.5\%
        & 76.2\%
        & 36.3\% & 17.8\% & 25.2\%
        & 55.8\% \\
        & \textbf{MoEGen}
        & \textbf{90.5\%}
        & 34.7\% & 14.3\% & 32.0\%
        & \textbf{76.8\%}
        & \textbf{53.9\%} & \textbf{36.4\%} & \textbf{42.4\%}
        & \textbf{60.4\%} \\
        \midrule

        \multirow{4}{*}{LLaMA-2-13B}
        & LoRA
        & 89.1\%
        & \textbf{35.6\%} & 13.4\% & 31.6\%
        & 74.4\%
        & 27.9\% & 10.2\% & 18.5\%
        & 53.4\% \\
        & MoELoRA
        & 88.8\%
        & 35.1\% & 14.3\% & \textbf{32.3\%}
        & 75.2\%
        & 28.3\% & 11.1\% & 18.8\%
        & 53.8\% \\
        & MoLA
        & 88.9\%
        & 35.5\% & \textbf{14.6\%} & 31.9\%
        & 75.4\%
        & 29.0\% & 11.1\% & 18.8\%
        & 53.8\% \\
        & \textbf{MoEGen}
        & \textbf{89.5\%}
        & 34.8\% & 14.1\% & 31.1\%
        & \textbf{75.8\%}
        & \textbf{50.0\%} & \textbf{30.8\%} & \textbf{38.6\%}
        & \textbf{58.8\%} \\
        \bottomrule
      \end{tabular}
    }
  \end{table*}

\begin{wraptable}{r}{0.55\textwidth}
\small
\setlength{\tabcolsep}{3pt}
\renewcommand{\arraystretch}{1.05}
\begin{tabular}{@{}lccccc@{}}
\toprule
Method
& MedNLI
& BillSum
& HQS
& PubMedQA
& Avg. \\
& Acc.
& R-L
& R-L
& Acc.
& \\
\midrule
Top-1
& 0.8924
& 0.3838
& \textbf{0.3154}
& 0.7460
& 0.5844 \\

Top-2 + shared
& 0.8924
& 0.3824
& 0.3138
& 0.7420
& 0.5827 \\

Top-2
& \textbf{0.8952}
& \textbf{0.3861}
& 0.3113
& \textbf{0.7580}
& \textbf{0.5877} \\
\bottomrule
\end{tabular}

\caption{Ablation study on routing design choices.}
\label{tab:ablation_topk_shared}
\end{wraptable}
Adding an always-on shared expert does not improve performance. Compared with
standard top-2 routing, the shared-expert variant reduces the average score from
$0.5877$ to $0.5827$, with drops on BillSum, HQS, and PubMedQA. This suggests
that, under a fixed parameter budget, the shared expert may absorb part of the
learning signal and weaken the specialization of routed experts. Therefore, we
use top-2 routing without a shared expert as the default configuration.
\begin{figure*}[t]
    \centering
    \includegraphics[width=\textwidth]{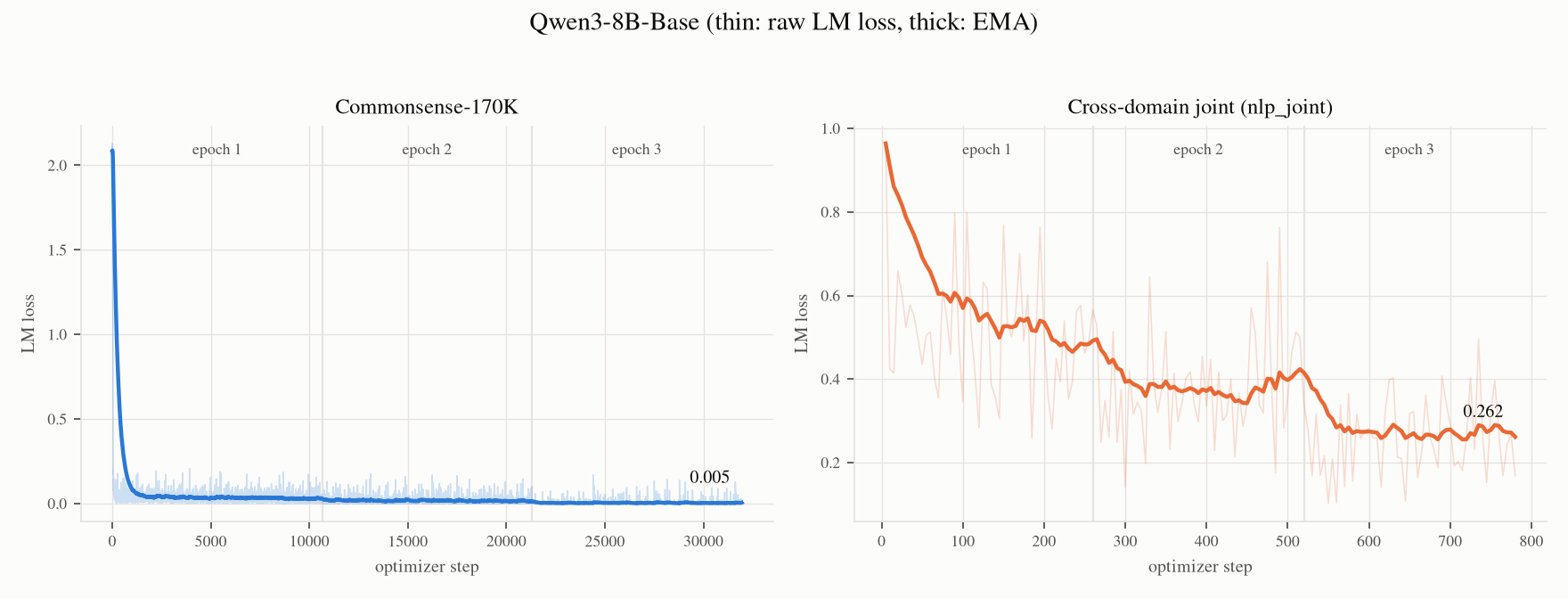}
    
    \caption{
Training Loss Example.
}
    \label{fig:MoEGen}
\end{figure*}

\section{Additional Qualitative Study}

We provide two additional BillSum predictions in
Table~\ref{tab:qual_app_billsum}, drawn from the same joint cross-domain
training run as Table~\ref{tab:qual_billsum}.  Table~\ref{tab:qual_billsum} presents a representative example from BillSum.
The reference summary describes the bill by naming the act and outlining its
main provisions. In contrast, LoRA, MoE-LoRA, and MoLA generate short legal
fragments copied from the bill text, failing to capture the main actions of the
legislation. The outputs of MoE-LoRA and MoLA are also highly similar,
suggesting that simply adding MoE-style adapters may still struggle to model
diverse task behaviors under joint multi-task adaptation.
MoEGen produces a more complete summary. It correctly
identifies the bill title and recovers the key legislative actions, including
the repeal of the excise tax and changes to fuel-related taxes. As a result,
its prediction is much closer to the reference summary and achieves a
substantially higher ROUGE-L score.

The two examples illustrate
distinct failure modes of the static LoRA-based baselines on legal
long-document inputs: collapse into a repetitive ``subsection X does not
apply'' loop in Example 1, and emission of mid-clause rules-of-the-House
boilerplate in Example 2. In both cases MoEGen opens with the act
name followed by ``Amends \ldots\ to\ldots'', the canonical BillSum
register, and enumerates the substantive actions present in the
reference. None of these behaviors require extra parameters:
MoEGen uses fewer trainable parameters than LoRA and MoLA and
roughly the same as MoE-LoRA in Table~\ref{tab:nlp_joint_full}, so the
difference is attributable to the input-conditioned parameterization
rather than to capacity. In particular, MoLA uses more than twice as many
trainable parameters as MoEGen yet exhibits the same qualitative
failures as the other two static baselines on these inputs.

\begin{table}[!tbp]
\centering
\small
\setlength{\tabcolsep}{2.5pt}
\caption{Per-sample inference latency (s). Max Tok. denotes the maximum number of generated tokens.}
\label{tab:latency}
\begin{tabular}{lrrrrr}
\toprule
Task & Max Tok. & LoRA & MoE-LoRA & MoLA & Ours \\
\midrule
MedNLI   & 8   & 0.058 & 0.062 & 0.074 & 0.090 \\
HQS      & 64  & 0.309 & 0.336 & 0.437 & 0.337 \\
PubMedQA & 256 & 0.283 & 0.324 & 0.798 & 0.306 \\
BillSum  & 384 & 0.966 & 1.025 & 2.259 & 1.006 \\
\bottomrule
\end{tabular}
\vspace{-4mm}
\end{table}
\label{qual_study_additional}

\section{Inference Cost}

We further evaluate the inference cost of MoEGen on the cross-domain
joint benchmark. Table \ref{tab:latency} reports the per-sample latency of each
method using LLaMA-2-13B on a single B200 GPU with batch size 32.
MoEGen introduces an input-dependent routing and generation cost, but
the overhead remains small on longer-generation tasks. The largest relative
gap appears on MedNLI, where the output is very short, and the fixed routing
cost is less amortized. As the generation length increases, this gap becomes
much smaller: MoEGen is only 9.1\% slower than LoRA on HQS, 8.1\%
slower on PubMedQA, and 4.1\% slower on BillSum. It also remains
substantially faster than MoLA on the longer-generation tasks.

\end{document}